\documentclass[letterpaper, 10 pt, journal, twoside]{IEEEtran}

\usepackage{amsmath,amsfonts}
\usepackage{algorithmic}
\usepackage{algorithm}
\usepackage{array}
\usepackage[caption=false,font=normalsize,labelfont=sf,textfont=sf]{subfig}
\usepackage{textcomp}
\usepackage{stfloats}
\usepackage{url}
\usepackage{verbatim}
\usepackage{graphicx}
\usepackage{cite}

\usepackage{bm}         
\usepackage{mathptmx} 
\usepackage{times}    
\usepackage{amsmath} 
\usepackage{amssymb}  
\usepackage[colorlinks=false, linkcolor=black, citecolor=black, urlcolor=black, pdfborder={0 0 0}]{hyperref}     
\usepackage{array}
\usepackage{multirow}
\usepackage[normalem]{ulem}     
\useunder{\uline}{\ul}{}     
\hypersetup{pdfborder={0 0 0}}
\usepackage{url}
\usepackage{hyperref}
\usepackage{xurl}
\usepackage{tabularx}
\usepackage{makecell}
\usepackage{xcolor}

\begin{document}

\title{Vibration-aware Lidar-Inertial Odometry based on \\
Point-wise Post-Undistortion Uncertainty}

\author{Yan Dong$^{1}$, Enci Xu$^{1}$, Shaoqiang Qiu$^{1}$, Wenxuan Li$^{1}$, Yang Liu$^{2}$ and Bin Han$^{1*}$

\thanks{Manuscript received 20 March 2025; accepted 30 June 2025. Date of publication DAY MONTH YEAR; date of current version DAY MONTH YEAR. This letter was recommended for publication by Associate Editor XXX and Editor XXX upon evaluation of the reviewers' comments. This work was supported in part by the National Natural Science Foundation of China under Grant 52375015, and in part by the Interdisciplinary Research Program (Robotics and Artificial Intelligence) of HUST under Grant 2024JCYJ037.}

\thanks{$^{1}$Yan Dong, Enci Xu, Shaoqiang Qiu, Wenxuan Li, and Bin Han are with the State Key Laboratory of Intelligent Manufacturing Equipment and Technology, School of Mechanical Science and Engineering, Huazhong University of Science and Technology, Wuhan, China. (e-mail: \tt\footnotesize binhan@hust.edu.cn)}%

\thanks{$^{2}$Yang Liu is with School of Artificial Intelligence and Automation, Huazhong University of Science and Technology, Wuhan, China.}%

\thanks{Digital Object Identifier (DOI): see top of this page.}
}

\markboth{IEEE Robotics and Automation Letters. Preprint Version. July, 2025.}
{Dong \MakeLowercase{\textit{et al.}}: Point-wise Post-Undistortion Uncertainty.}

\maketitle

\begin{abstract}

High-speed ground robots moving on unstructured terrains generate intense high-frequency vibrations, leading to LiDAR scan distortions in Lidar-inertial odometry (LIO). Accurate and efficient undistortion is extremely challenging due to (1) rapid and non-smooth state changes during intense vibrations and (2) unpredictable IMU noise coupled with a limited IMU sampling frequency. To address this issue, this paper introduces post-undistortion uncertainty. First, we model the undistortion errors caused by linear and angular vibrations and assign post-undistortion uncertainty to each point. We then leverage this uncertainty to guide point-to-map matching, compute uncertainty-aware residuals, and update the odometry states using an iterated Kalman filter. We conduct vibration-platform and mobile-platform experiments on multiple public datasets as well as our own recordings, demonstrating that our method achieves better performance than other methods when LiDAR undergoes intense vibration.

\end{abstract}


\section{Introduction}

\IEEEPARstart{I}{n} recent years, field robots have developed rapidly \cite{Darpa-Review, Luliang}, and simultaneous localization and mapping (SLAM) technology serves as the foundation for achieving full autonomy \cite{Darpa-Cerberus}. Ground robots—whether wheeled, legged, or tracked—inevitably interact with the terrain, making them susceptible to the effects of uneven surfaces. These external stimuli further induce intense vibrations in the robot’s body, adversely impacting LiDAR-inertial odometry (LIO).

Vibration induces motion distortion in LiDAR scans. The process of scan distortion removal, commonly referred to as “motion compensation,” “de-skewing” or “undistortion”, has become a standard procedure in many LIO systems and has been widely studied \cite{LOAM, Fastlio2, LIO-SAM, Coco-LIC}. Most of these undistortion methods rely on IMU data or trajectory smoothing assumptions to estimate motion within a scan and subsequently correct the positions of the scanned points. Unfortunately, under intense vibration, the state changes more rapidly and non-smoothly, leading to unpredictable IMU noise. While accuracy can be improved to some extent by employing vibration isolation, high-performance IMUs, or more complex computational methods \cite{TIM}, these approaches inevitably increase hardware and computational costs. Therefore, achieving accurate undistortion in high-vibration scenarios remains challenging within acceptable hardware constraints and computational time limits.

\begin{figure}[t]
  \centering
  \includegraphics[width=0.9\linewidth]{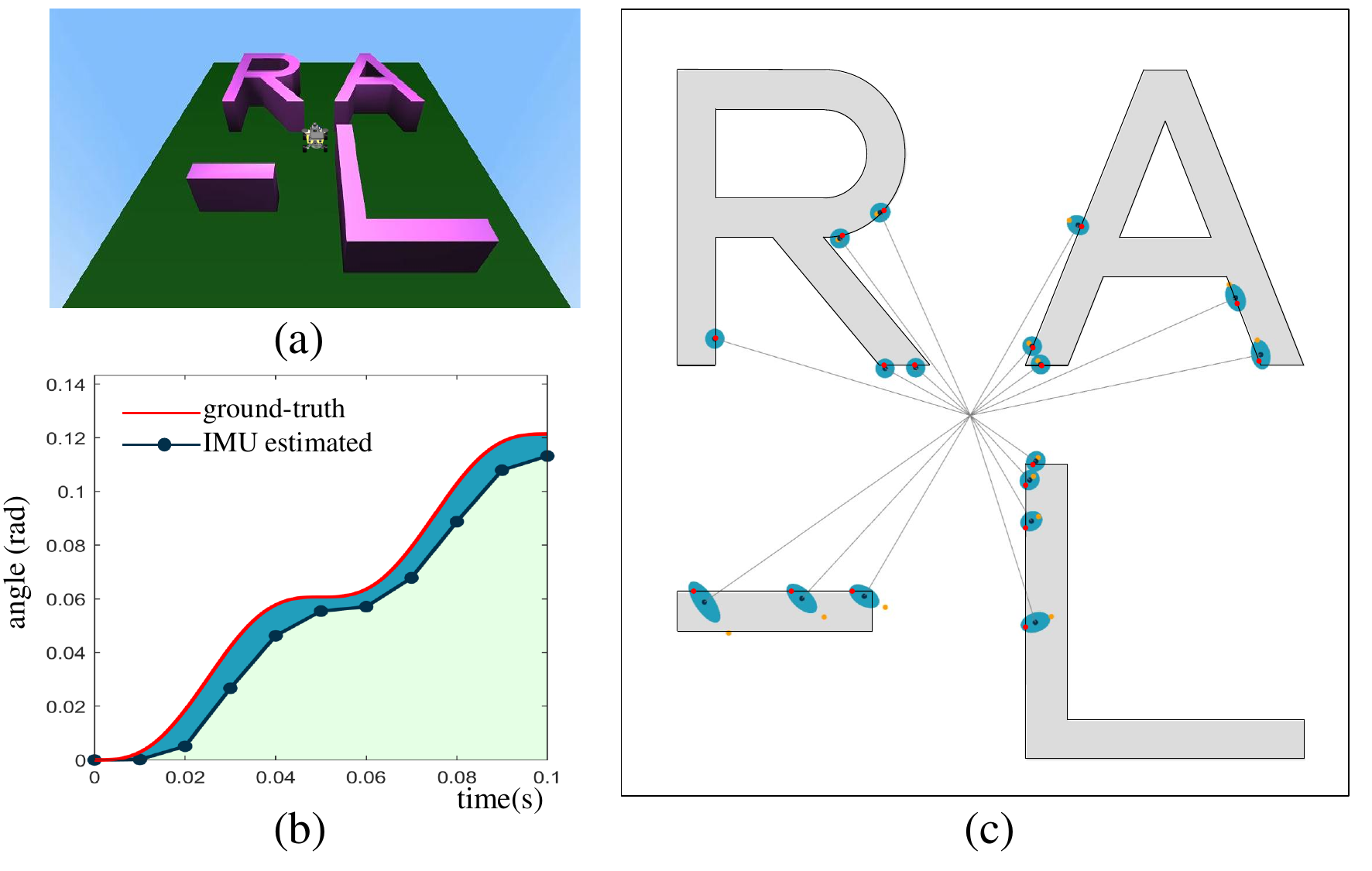}
  \caption{Illustration of the proposed post-undistortion uncertainty. (a) A high-frequency angular vibration is applied to the robot equipped with LiDAR and IMU in the simulation engine. (b) Ground-truth poses from the simulation engine vs. estimated poses from a 100Hz noisy IMU during a single LiDAR scan (0.1s). The gap between the two curves (black region) represents the undistortion error. (c) Yellow, black, and red dots denote raw points, IMU-aided undistorted points, and error-free undistorted points obtained from ground-truth poses, respectively. black ellipsoids represent the proposed post-undistortion uncertainties.}
  \label{Fig:explanation}
\end{figure}

We argue that point cloud undistortion does not need to be highly accurate—as long as the uncertainty after undistortion can be estimated. Therefore, we introduce “point-wise post-undistortion uncertainty” to quantify the uncertainty of each point after undistortion under intense vibration (Fig.~\ref{Fig:explanation}). This paper aims to address two key challenges: 1) \textit{How to represent the errors introduced by vibration after point cloud undistortion}, and 2) \textit{How to leverage this uncertainty to enhance the LIO system}. The main contributions of this paper are:

\begin{figure*}[ht]
  \centering
  \includegraphics[width=\textwidth]{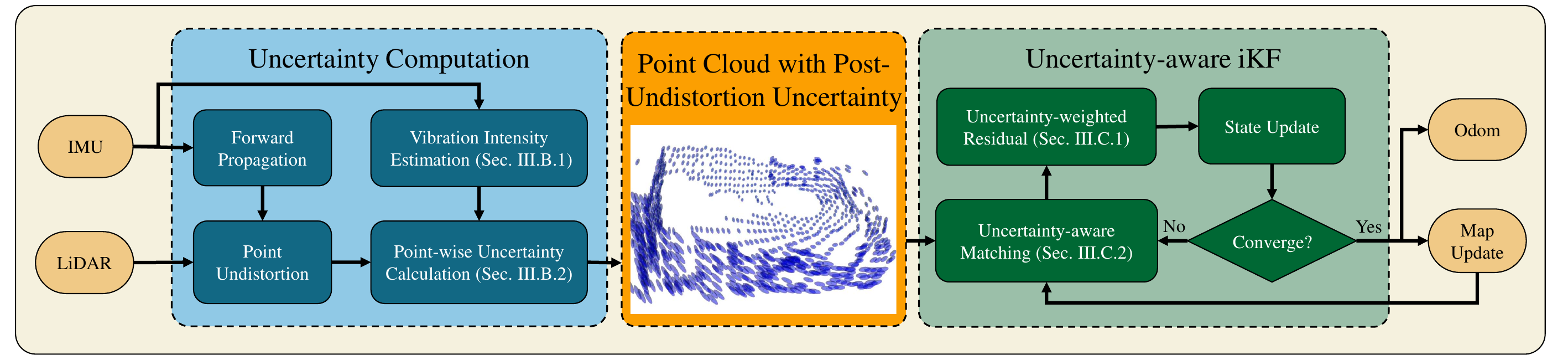}
  \caption{Overview of the proposed LIO system. The core of this system is that each point in a scan is assigned a point-wise post-undistortion uncertainty.}
  \label{Fig:method}
\end{figure*}

\begin{itemize}
\item We analyze the causes of undistortion errors from angular and linear vibrations and establish uncertainties for both rotation and translation, enabling the modeling of post-undistortion uncertainty for each scan point.

\item We propose a vibration-aware LIO framework based on this uncertainty, incorporating uncertainty-guided scan-to-map matching and uncertainty-weighted residual computation within an iterated Kalman filter (iKF) for state estimation.

\item We conduct both vibration-platform and mobile-platform experiments, demonstrating the effectiveness of our method. Additionally, we collect several challenging sequences using a wheeled robot in an unstructured environment. These sequences includes cm-level ground-truth trajectories and are available\footnote{\url{https://drive.google.com/drive/folders/1wtDIPiUBqs9FbEoeLWOyvwkLmb2QpORX}}.

\end{itemize}


\section{Related Work}
\subsection{Point Cloud Undistortion}

LiDAR requires a certain amount of time to complete a full scan, typically 100 ms, and estimating LiDAR motion during this period is crucial for undistortion. Early methods, such as LOAM \cite{LOAM}, estimate the motion of the current scan based on historical data, making them suitable only for slow-motion scenarios. By incorporating higher-frequency sensors like IMUs and wheel odometry, linear interpolation and integration have been widely adopted to estimate motion during the scan \cite{Fastlio2, LIO-SAM}. To handle rapidly changing motions, some approaches divide a single scan into smaller segments for more fine-grained undistortion \cite{Ensemble}. However, segmentation alone cannot fundamentally eliminate errors.

Many researchers have adopted a continuous-time (CT) approach for undistortion. The most commonly used methods employ B-spline interpolation to generate a smoother trajectory, utilizing either evenly spaced \cite{CLINS} or non-uniform \cite{Coco-LIC} control points to model LiDAR motion. However, in cases of aggressive motion, trajectory continuity may be violated. To address this, CT-ICP \cite{CT-ICP} introduces different constraints for intra-frame and inter-frame motions to achieve elastic undistortion. In addition to B-spline interpolation, the Gaussian Processes (GP) has also been applied \cite{TIM}. However, CT-based methods generally involve significant computational overhead, making real-time performance challenging to achieve.

\subsection{Point Uncertainty Modeling}

LiDAR emits and receives laser pulses to obtain distance information. Consequently, the measurements inherently involve geometric uncertainty. Since a LiDAR beam may exhibit errors in both angle and range, it is common to model range errors and beam divergence as Gaussian distributions \cite{LiDAR-Noise1, LiDAR-Noise2}. Recently, CTA-LO \cite{CTA-LO} has considered the effects of surface reflectance and developed a spot error entropy model. Despite the variety of modeling strategies, most SLAM algorithms simply model measurement uncertainty as a 3D Gaussian distribution in space \cite{Fastlio2, PLIO}.

Researchers have also explored how to utilize point uncertainty in scan registration to minimize the impact of measurement errors. In \cite{DistortionICP}, the effect of point skewing on ICP is analyzed, and the pipeline directly incorporates skewed points without performing undistortion. In \cite{CRV21}, various weighting methods are proposed to address measurement errors.



\section{Method}    \label{sec:3}

The proposed algorithm is illustrated in Fig.\ref{Fig:method}. The algorithm primarily consists of two parts: 1) obtaining the point-wise post-undistortion uncertainty, and 2) updating the estimate state by an iKF considering this uncertainty. In this section, we first introduce the notation for state estimation in Sec.\ref{sec:3.1}. 

\subsection{State Definition and Iterated Kalman Filter}    \label{sec:3.1}

The state definition and on-manifold computation in this section are similar to those in Fast-LIO \cite{Fastlio2}. The odometry coordinate frame is aligned with the IMU on the robot, denoted as $I$. The global frame $G$ is defined as the IMU's coordinate system after the system initialization. The entire LIO framework estimates the robot's pose in the global frame at any given time. The complete state is defined as:
\begin{equation}  \label{eq:1}
    \bm{x}  \triangleq [^{G}\bm{R}_I^T \quad ^{G}\bm{p}_I^T \quad ^{G}\bm{v}_I^T \quad \bm{b}_\omega^T \quad \bm{b}_a^T]^T
\end{equation}
\noindent where $^{G}\bm{R}_I$ and $^{G}\bm{p}_I$ represent the rotation and translation from the IMU frame to the global frame, $^{G}\bm{v}_I$ denotes the velocity, and $\bm{b}_\omega$ and $\bm{b}_a$ are the IMU random walk biases.

When an IMU measurement comes, the state $\bm{x}$ transits by the discrete transition model:
\begin{equation}  \label{eq:2}
    \bm{x}_{k+1} = \bm{x}_k \boxplus (\Delta t f(\bm{x}_i, \bm{u}_i, \bm{w}_i))
\end{equation}

The input $\bm{u}$, process noise $\bm{w}$ and function $f$ are defined as:
\begin{equation}  \label{eq:3}
\scalebox{0.9}{$
    f(\bm{x},\bm{u},\bm{w})=\begin{bmatrix}
                \bm{\omega}_m- {\color{black}\bm{b}_\omega-\bm{n}_\omega} \\
                ^{G}\bm{v}_I + \frac{1}{2}(^{G}\bm{R}_I(\bm{a}_m-\bm{b}_a-\bm{n}_a) + \bm{g})\Delta t \\
                ^{G}\bm{R}_I(\bm{a}_m-\bm{b}_a-\bm{n}_a) + \bm{g} \\
                {\color{black}\bm{n}_{b_\omega}} \\
                \bm{n}_{b_a}
            \end{bmatrix}
            $}
\end{equation}
\noindent where $\bm{\omega}_m$ and $\bm{a}_m$ are the raw IMU measurements, $\bm{n}_\omega$, $\bm{n}_a$, $\bm{n}_{b_\omega}$ and $\bm{n}_{b_a}$ are measurement noises in $\bm{\omega}_m$, $\bm{a}_m$, $\bm{b}_\omega$ and $\bm{b}_a$, respectively, which are modeled as random walk. $\bm{g}$ is the gravity.

\begin{figure}[t]
  \centering
  \includegraphics[width=0.85\linewidth]{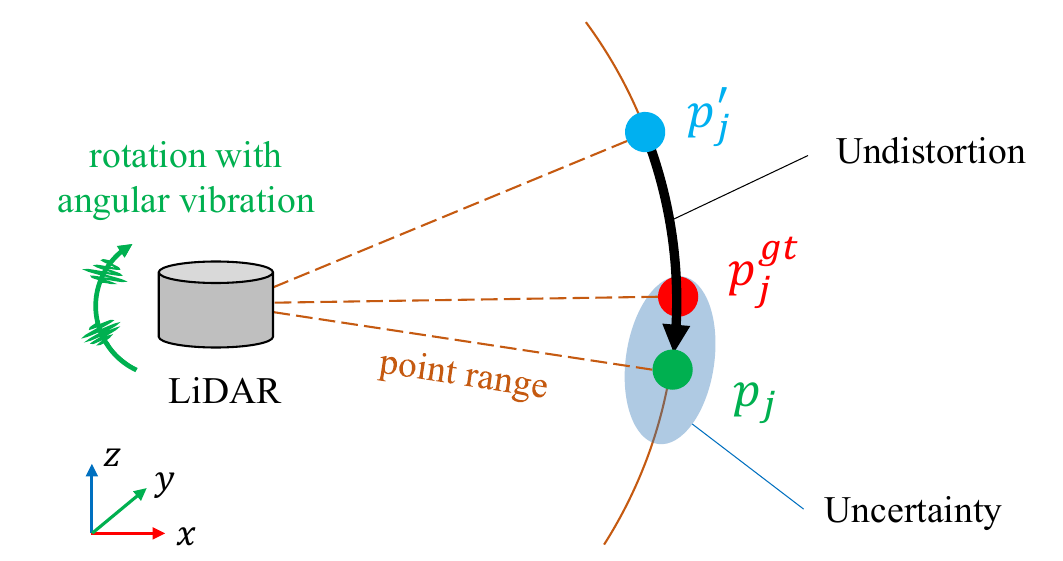}
  \caption{Post-undistortion uncertainty caused by rotational vibration. During a single scan, a sensor experiences rotational jitter along the $y$ axis. A scan point $\bm{p}_j^{'}$ is undistorted using IMU data resulting in $\bm{p}_j$, which deviates from the ideal undistorted point $\bm{p}_j^{gt}$. The rotational error caused by angular vibrations contributes to the uncertainty of $\bm{p}_j$.}
  \label{Fig:ErrorAnalyze}
\end{figure}

\renewcommand{\arraystretch}{1.5}
\begin{table}[t]    
\caption{Notations used in proposed method} \label{Tab0:notation}
\begin{tabular}{|>{\centering\arraybackslash}m{0.8cm}|p{6.8cm}|} 
\hline
Symbol &  \multicolumn{1}{c|}{Description} \\ \hline
$^{L}\bm{p}'_j$ & raw point, the $j$-th point in a single LiDAR scan before the undistortion process   \\    \hline
$^{L}\bm{p}_j$ & calculated undistort point, the $j$-th point position after undistortion based on LiDAR motion estimated by the IMU    \\  \hline
$^{L}\bm{p}^{gt}$ & ground-truth point, the accurate position of the $j$-th point {\color{black}without measurement noise} after undistortion by the ideal error-free LiDAR motion \\ \hline
\end{tabular}
\end{table}
\renewcommand{\arraystretch}{1.0}

\subsection{Point-wise Post-Undistortion Uncertainty}   \label{sec:3.2}
\subsubsection{Undistortion Error Analysis}     \label{sec:3.2.1}
For clarity, we define the notations used in this section in Table \ref{Tab0:notation}. All variables are expressed in the LiDAR coordinate frame.

When the robot undergoes violent vibrations, the IMU generates significant measurement noise. Consequently, even if the LiDAR itself has no measurement error, undistorting the point cloud using these IMU measurements does not strictly correct the points to their true positions. Figure \ref{Fig:ErrorAnalyze} presents a demonstration case.

We revisit the undistortion process: all points in a scan are aligned to the initial time $t_0$ of the scan using IMU data. In the absence of errors, the following condition should hold: 
\begin{equation} \label{eq:4}
    ^{L}\bm{p}_j = \bm{R}^{t_0}_j\ ^{L}\bm{p}^{'}_j\ + \bm{T}^{t_0}_j\ 
\end{equation}

\noindent where $\bm{R}_j^{t_0}$ and $\bm{T}_j^{t_0}$ represent the rotation and translation from the initial time $t_0$ of the scan to the $j$-th scan point, computed via IMU integration. Under intense vibration, due to IMU measurement noise and discretization issues, the computed $\bm{R}_j^{t_0}$ and $\bm{T}_j^{t_0}$ deviate from the error-free rotation and translation. We denote the ground-truth rotation and translation as $\bm{R}^{gt}$ and $\bm{T}^{gt}$, respectively. {\color{black}$\bm{n_p}$ is introduced as the error between $\bm{p}^{gt}$ and $\bm{p}$. We omit the coordinate system $L$, the subscripts $j$ and $t_0$ for simplicity.} The relationship between the ground-truth point $\bm{p}^{gt}$ and the computed undistorted point $\bm{p}$ is:{\color{black}
\begin{equation} \label{eq:5} 
\begin{split}
     \bm{p}^{gt} =& \bm{R}^{gt} \, (\bm{p}'-\bm{n_p}) + \bm{T}^{gt} \\
     =& (\bm{R}^{gt} \bm{R}^{-1})\bm{p} + (\bm{T}^{gt} - (\bm{R}^{gt} \bm{R}^{-1})\bm{T}) - \bm{R}^{gt}{\bm{n_p}} \\
       \triangleq& \delta\!\bm{R} \cdot \bm{p} +\delta\!\bm{T} - \delta\!\bm{R} \cdot \bm{R} \cdot {\bm{n_p}}
\end{split}
\end{equation} }
where $\delta \bm{R}$ and $\delta \bm{T}$ are the rotational and translational undistortion errors.

We can calculate $\bm{R}, \bm{T}$ but cannot accurately determine $\delta\!\bm{R}, \delta\!\bm{T}$, just as we cannot directly obtain $\bm{R}^{gt}, \bm{T}^{gt}$. Therefore, we characterize the error between the $\bm{p}^{gt}$ and $\bm{p}$ using the uncertainty associated with $\bm{p}$.

\subsubsection{Point-wise Post-Uncertainty Calculation}      \label{sec:3.3.2}
To facilitate understanding, we first analyze angular errors. During the undistortion process, we denote the error between the computed and true rotations along the $x$, $y$, and $z$ axes as $\bm{\delta}_r=[\delta_{rx},\delta_{ry},\delta_{rz}]^T$. Given that the error is small, the rotation matrix corresponding to the errors along different axes can be approximated as:
\begin{equation} \label{eq:6} 
\scalebox{0.7}{$
    \delta \bm{R}_x \approx\
    \begin{bmatrix}
        1 & 0 & 0 \\
        0 & 1 & -\delta_{rx} \\
        0 & \delta_{rx} & 1 
    \end{bmatrix},
    \delta \bm{R}_y \approx\
    \begin{bmatrix}
        1 & 0 & \delta_{ry} \\
        0 & 1 & 0 \\
        -{\color{black}\delta_{ry}} & 0 & 1
    \end{bmatrix},
    \delta \bm{R}_z \approx\ 
    \begin{bmatrix}
        1 & -\delta_{rz} & 0 \\
        \delta_{rz} & 1 & 0 \\
        0 & 0 & 1
    \end{bmatrix}
$}
\end{equation}

Higher-order terms can be ignored, allowing the order of rotations to be considered irrelevant. Consequently, the final rotation matrix can be expressed as:
\begin{equation} \label{eq:7} 
\scalebox{0.9}{$
\delta \bm{R} \approx\ 
    \begin{bmatrix}
        1 & -\delta_{rz} & \delta_{ry} \\
        \delta_{rz} & 1 & -\delta_{rx} \\
        -\delta_{ry} & \delta_{rx} & 1
    \end{bmatrix}
$}
\end{equation}

We assume that the angular error $\bm{\delta_r}$ follows a Gaussian distribution. Ideally, the mean of the error is zero: 
\begin{equation}  \label{eq:9} 
\bm{\delta}_r \sim \ N(0, \bm{\Sigma}_r), \quad \bm{\Sigma}_r=diag\{\sigma_{rx}^2,\sigma_{ry}^2,\sigma_{rz}^2\}
\end{equation}

\noindent where $diag\{\cdot\}$ represents a diagonal matrix. 

The uncertainty of $\bm{p}$ represents the probability of $\bm{p}^{gt}$ appearance, which depends on the rotational calculation error $\delta_r$. To analyze the influence of $\bm{\delta}_r$, the derivative of $\bm{p}^{gt}$ with respect to $\bm{\delta}_r$ is calculated:

\begin{equation}  \label{eq:10} 
\scalebox{0.95}{$
\frac{\partial \bm{p}^{gt}}{\partial \bm{\delta}_r }= 
    \begin{bmatrix}
        0 & z & -y \\
       -z & 0 & x \\
        y & -x & 0
    \end{bmatrix} = {\color{black}{\lfloor \bm{p} \rfloor}_\times}
$}
\end{equation}
{\color{black}
where ${\lfloor \cdot \rfloor}_\times$ is skew-symmetric of a vector.

Therefore, the uncertainty of the calculated undistorted point $\bm{p}$ due to rotational error is:
\begin{equation}  \label{eq:11} 
\bm{\Sigma}_p^{rot} = {\lfloor \bm{p} \rfloor}_\times \, \bm{\Sigma}_r \, {\lfloor \bm{p} \rfloor}_\times^T = {\lfloor \bm{p} \rfloor}_\times \cdot diag\{\sigma_{rx}^2,\sigma_{ry}^2,\sigma_{rz}^2\} \cdot {\lfloor \bm{p} \rfloor}_\times^T
\end{equation}
}
Now we address the translational uncertainty in a similar manner. We assume that the translational error $\bm{\delta T} = [\delta_{Tx},\delta_{Ty},\delta_{Tz}]^T \sim N(0, \bm{\Sigma}_T)$, where $\bm{\Sigma}_T = diag\{\sigma^2_{Tx},\sigma^2_{Ty}, \sigma_{Tz}^2\}$. Since the translational uncertainty is independent of the point position, it does not involve scaling or stretching:
\begin{equation} \label{eq:12} 
\bm{\Sigma}_p^{trans} = \bm{\Sigma}_T 
= diag\{\sigma^2_{Tx},\sigma^2_{Ty},\sigma^2_{Tz} \}
\end{equation} 

Now we deal with $\bm{n_p}$, the LiDAR's own measurement error. Based on \cite{RAL22VoxelMap}, the LiDAR point's bearing angle is $\bm{\phi} \in \bm{S}^2$ and measured range is $d$, and assume their noise components follow Gaussian distributions: $\bm{\delta}_\phi \sim N(0, \text{diag}\{\sigma_\phi, \sigma_\phi\})$ and $\delta_d \sim N(0, \sigma_d)$. Then we have $\bm{n_p} \sim N(0, \bm{\Sigma}_p^{meas})$, and:
\begin{align} \label{eq:13} 
\begin{split}
\bm{\Sigma}_p^{meas} &= \bm{A}_p \, \text{diag}\{\sigma_d, \sigma_\phi, \sigma_\phi\} \bm{A}_p^T \\
\bm{A}_p &= [\bm{\phi}, -d {\left\lfloor \bm{\phi} \right\rfloor}_\times O(\bm{\phi})]
\end{split}
\end{align}
\noindent where $O(\phi)=[O_1, O_2]$ is an orthonormal basis of the tangent plane at $\bm{\phi}$.

{\color{black}
We rewrite Equation (\ref{eq:5}) as $\bm{p}^{gt}= \bm{p} + \delta \bm{p}$, where $\delta \bm{p}$ is the error between $\bm{p}^{gt}$ and $\bm{p}$:
\begin{equation} \label{eq:a1} 
\begin{split}
\delta \bm{p} \triangleq& \bm{p}^{gt} - \bm{p} = {\lfloor \bm{\delta_r} \rfloor}_\times \bm{p} + \delta \bm{T} - (\bm{I}+ {\lfloor \bm{\delta_r} \rfloor}_\times) \bm{R} \cdot \bm{n_p} \\
=& -{\lfloor \bm{p} \rfloor}_\times \bm{\delta_r} + \delta \bm{T} - \bm{R} \cdot \bm{n_p} - {\lfloor \bm{\delta_r} \rfloor}_\times \bm{R} \cdot \bm{n_p}
\end{split}
\end{equation}
Since $\bm{\delta_r}$ and $\bm{n_p}$ are small values, the last term can be neglected. The $\bm{\delta_r}, \delta\bm{T}$ and $\bm{n_p}$ obey Gaussian distributions by Equation (\ref{eq:11}-\ref{eq:13}), so the distribution of $\delta \bm{p}$ is $N(0, ^L \bm{\Sigma}_p)$, which represents the uncertainty of $\bm{p}$:
\begin{align}  \label{eq:14} 
\begin{split}
^L\bm{\Sigma}_p   &=    \bm{\Sigma}_p^{rot} + \bm{\Sigma}_p^{trans} + \bm{R} \, \bm{\Sigma}_p^{meas} \, \bm{R}^T
\end{split}
\end{align}
where $\bm{R}$ here is $\bm{R}_{j}^{t_0}$ in Equation (\ref{eq:4}), which is the estimated undistortion rotation of current point.
}

\begin{figure}[t]
  \centering
  \includegraphics[width=0.9\linewidth]{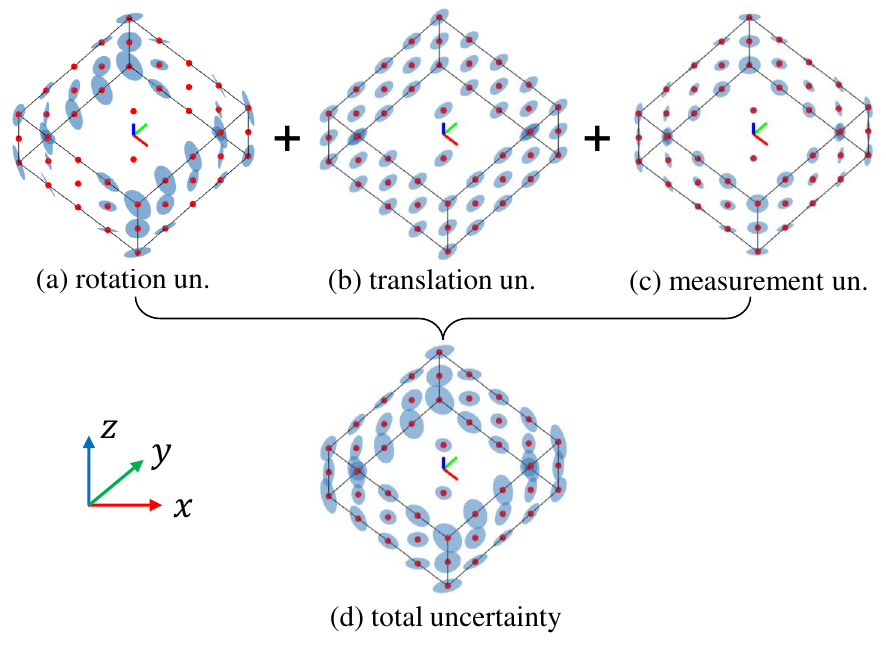}
  \caption{Different types of uncertainty. This figure shows the uncertainty caused by angular vibration along the $y$ and $z$ axes (a) and linear vibration along the $y$-axis (b). (c) and (d) are the measurement and total uncertainty.}
  \label{Fig4:4Uncertainty}
\end{figure}

{\color{black}
Figure \ref{Fig4:4Uncertainty} provides a visual representation of these uncertainties.} In practice, the measurement uncertainty $\bm{\Sigma}_p^{meas}$ is typically provided by the LiDAR manufacturer or obtained through manual calibration. The remaining question is how to compute $[\sigma_{rx},\sigma_{ry},\sigma_{rz}]$ for $\bm{\Sigma}_p^{rot}$ and $[\sigma_{Tx},\sigma_{Ty},\sigma_{Tz}]$ for $\bm{\Sigma}_p^{trans}$.

\subsubsection{Vibration Intensity Estimation }  \label{sec:3.2.3}

More intense vibrations result in larger rotational and translational errors, as the angular velocity provided by the IMU and the velocity estimated from state estimation contain significant inaccuracies. Therefore, we assess vibration intensity to estimate the error for undistortion.

Angular vibration is directly evaluated using IMU data. During one scan, $M$ IMU data points are collected, with gyroscope measurements recorded as $\{\bm{\omega}_I(i), i=1...M\}$, {\color{black} and estimated linear velocity $\{^G\bm{v}_I(i), i=1...M\}$ from Equatoin (\ref{eq:1})}. Given the LiDAR-IMU extrinsics $^{I}\bm{T}_L=[^{I}\bm{R}_L,^{I}\bm{t}_L]$, the LiDAR's angular and linear velocity are computed as:
\begin{equation} \label{eq:15}
    \bm{\omega}_L(i)= (^{I}\bm{R}_L)^T \bm{\omega}_I(i), 
    \quad 
    {\color{black} \bm{v}_L(i) = (^{I}\bm{R}_L)^T ({^I\bm{R}_G}) ^{G}\bm{v}_I(i) }
\end{equation}

{\color{black}
Then average value calculated as $\bm{\omega}_{ave}, \bm{v}_{ave}$. The angular and linear vibration intensity denoted as $\bm{k}_\omega$ and $\bm{k}_v$ are heuristic defined by the mean absolute deviation (MAD) of the angular and linear velocity respectively:
\begin{equation} \label{eq:16}
    \bm{k}_\omega \triangleq \frac{1}{M} \sum_{i=1}^M |\bm{\omega}_L(i) - \bm{\omega}_{ave}|,
    \quad 
    \bm{k}_v \triangleq \frac{1}{M} \sum_{i=1}^M |\bm{v}_L(i) - \bm{v}_{ave}|,
\end{equation}
}

When performing undistortion, each point in the scan is aligned to the start of the scan, and we use $\delta t_{j0}$ to represent this time interval. Consequently, if $\bm{p}$ is close to the beginning, it is relatively less affected by vibration. We calculate the uncertainty in Equation (\ref{eq:10})(\ref{eq:12}) by:
\begin{equation}  \label{eq:19} 
\begin{split}
\bm{\sigma_r} = \gamma \, \delta t_{j0} \, \bm{k}_\omega, \quad
\bm{\sigma_T} = \gamma \, \delta t_{j0} \, \bm{k_v}
\end{split}
\end{equation}
\noindent where $\bm{\sigma_r} \triangleq [\sigma_{rx},\sigma_{ry},\sigma_{rz}]^T$ and $\bm{\sigma_T} \triangleq [\sigma_{Tx},\sigma_{Ty},\sigma_{Tz}]$. $\gamma$ is a hyperparameter that quantifies the uncertainty scale under specific vibration intensities. {\color{black} Higher IMU frequencies yield smaller errors, allowing for a smaller $\gamma$. We empirically set $\gamma=0.1$ for our experiments (for 100-200 Hz IMU). We observe that the proposed method is not sensitive to $\gamma$.}

\subsection{Uncertainty-Aware Iterated Kalman Filter} \label{sec:3.3}

Once the post-undistortion uncertainty is obtained, we employ an iKF framework to iteratively update the state until convergence. In each iteration, we perform vibration-aware matching for each LiDAR point and update the state based on the residuals computed from the observation model.

\subsubsection{Observation Model}   \label{sec:3.3.1}

The error between $^{L}\bm{p}^{gt}_j$ and the $^{L}\bm{p}_j$ is denoted as $^{L}\bm{n}_j$. According to Equation (\ref{eq:14}), $^{L}\bm{n}_j$ follows a Gaussian distribution with covariance $^{L}\bm{\Sigma}_p$. After the $\kappa$-th iteration of the $k$-th scan, we have: 
\begin{equation}  \label{eq:20} 
\begin{split}
^G\bm{p}_j^{gt} &= (^G\bm{\Hat{T}}_{I_k}^{\kappa}\ ^I\bm{T}_L)\ ^{L}\bm{p}_j^{gt} = ^{G}\bm{p}_j + ^{G}\bm{n}_j \\
^G\bm{\Sigma}_{pj} &= (^G\bm{\Hat{T}}_{I_k}^{\kappa}\ ^I\bm{T}_L) \, ^L\bm{\Sigma}_{\bm{p}_j}\ (^G\bm{\Hat{T}}_{I_k}^{\kappa}\ ^I\bm{T}_L)^T
\end{split}
\end{equation}
where $^{G}\bm{\Hat{T}}_{I_k}^\kappa = [^{G}\bm{\Hat{R}}_{I_k}^\kappa, ^{G}\bm{\Hat{p}}_{I_k}^\kappa]$ represents the robot's current pose, and $^I\bm{T}_L$ is the LiDAR-IMU extrinsics. 

After forward propagation \cite{Fastlio2} by setting the measurement noise $\bm{w}=\bm{0}$ in Equation (\ref{eq:3}). If the computation of $^{G}\bm{\Hat{T}}_{I_k}^\kappa$ is correct, $^{L}\bm{p}^{gt}_j$ should lie on the global map. We use the point-to-plane distance as the observation model:
\begin{equation} \label{eq:21} 
\begin{split}
  0 = h_j(\bm{\Hat{x}}_{k}^{\kappa},^G\bm{n}_j) &\triangleq \, ^{G}\bm{u}_j^T((^{G}\bm{p}_j + ^{G}\bm{n}_j)-^{G}\bm{q}_j)   \\ 
  ^G\bm{n}_j  &\sim N(0,\ ^G\bm{\Sigma}_{\bm{p}_j})  \\
\end{split}
\end{equation}
\noindent where $^{G}\bm{u}_j^T$ is the normal of the plane in the map corresponding to $^{G}\bm{p}_j$, and $^{G}\bm{q}_j$ is any point in that plane.

\begin{figure}[t]
  \centering
  \includegraphics[width=0.95\linewidth]{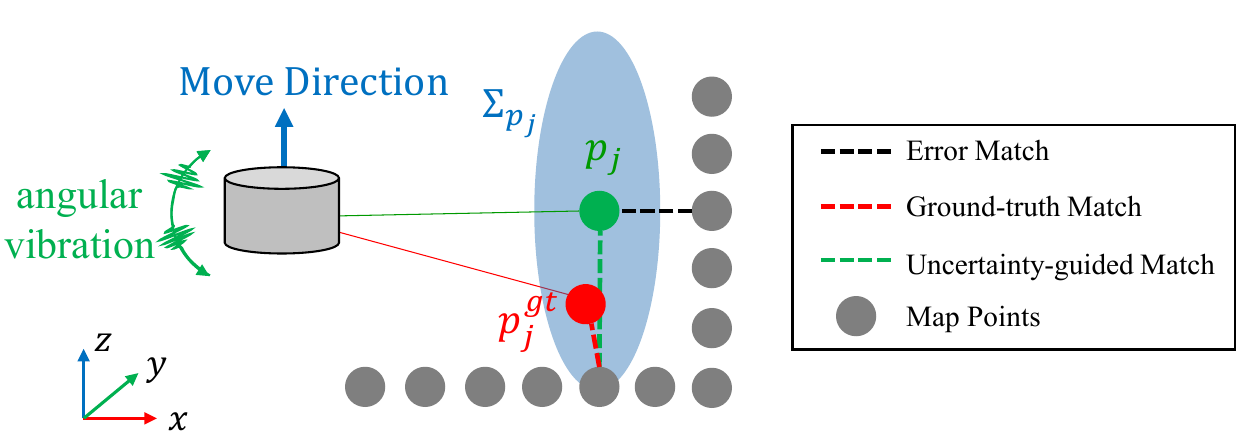}
  \caption{Demonstration of the uncertainty-guided matching process. When the LiDAR moves upward, the ground-truth point $\bm{p}_j^{gt}$ should be matched to the bottom of the map. However, due to undistortion errors, the calculated undistorted point $\bm{p}_j$ would be incorrectly matched to the right side of the map. By incorporating uncertainty, the uncertainty-guided matching process can still ensure a correct match.}
  \label{Fig:Match}
\end{figure}

\subsubsection{Uncertainty-Guided Scan-to-Map Matching} \label{sec:3.3.2b}

Based on the observation model, we need to find the “nearest" corresponding plane in the map for $\bm{p}_j$. However, due to undistortion errors caused by vibration, the nearest plane may not always be the correct one. To address this, we adopt an uncertainty-guided matching strategy (Fig.\ref{Fig:Match}).

Let $\mathbb{M}$ denote the set of all map points. We search for $K$ points from the entire map $\mathbb{M}$ to construct the point set $\mathbb{S}$ for plane fitting, ensuring that each point in $\mathbb{S}$ has a smaller Mahalanobis distance to $^{G}\bm{p}_j$ than other map points. Since each $^{G}\bm{p}_j$ has a unique covariance $^G\bm{\Sigma}_{\bm{p}_j}$, finding $K$ points from the entire map $\mathbb{M}$ incurs a significant computational burden. Common point cloud data structures, e.g., octree \cite{ioctree} and kd-tree \cite{ikdtree}, do not support nearest-neighbor search based on varying covariances of individual point. 

Therefore, we resort to an alternative approach. We first search for $K_c$ candidate points forming a set $\mathbb{S}_c$ ($K_c>K$, and we set $K_c=2K$ in our experiments). We then compute the Mahalanobis distance for each candidate and select the $K$ points with the smallest distances, forming the matching set $\mathbb{S}_m$ for $^{G}\bm{p}_j$:
\begin{equation} \label{eq:22} 
\begin{split}
\mathbb{S}_c &= \operatorname*{argmin}_{\mathbb{S} \subset \mathbb{M}, |\mathbb{S}|=K_c} (^G\bm{p}_j - \bm{m})^T (^G\bm{p}_j - \bm{m}), \quad \bm{m} \in \mathbb{M} \\
\mathbb{S}_m &= \operatorname*{argmin}_{\mathbb{S} \subset \mathbb{S}_c,|\mathbb{S_c}|=K} (^G\bm{p}_j - \bm{s})^T \, ^G\bm{\Sigma}_{\bm{p}_j}^{-1} \, (^G\bm{p}_j - \bm{s}), \quad \bm{s} \in \mathbb{S}_c
\end{split}
\end{equation}

\subsubsection{Uncertainty-Weighted Residual Computation} \label{sec:3.3.3}

After obtaining $\mathbb{S}_m$, we apply principal component analysis (PCA) to verify whether it satisfies the plane criteria. We then calculate the geometric center of $\mathbb{S}_m$ as $^{G}\bm{q}_j$ and the normal vector $^{G}\bm{u}_j$. The observation model in Equation (\ref{eq:21}) is linearized at $\bm{\Hat{x}}_k^\kappa$ (estimated state at the $\kappa$-th iteration in the $k$-th scan) for first-order approximation:
\begin{equation}  \label{eq:23} 
  0 \simeq h_j(\bm{\Hat{x}}_{k}^{\kappa},\ 0) + \bm{H}_j^{\kappa}\  \bm{\tilde{x}}_{k}^{\kappa} + r_j =z^{\kappa}_j + \bm{H}_j^{\kappa}\  \bm{\tilde{x}}_{k}^{\kappa} + r_j
\end{equation}

\noindent where $\bm{H_j}^{\kappa}$ is the Jacobian matrix of the measurement model $h_j$ with respect to error state $\bm{\tilde{x}}_{k}^{\kappa}$ (see \cite{Fastlio2} for details), and $\bm{z}_j^\kappa$ is the residual:
\begin{equation}  \label{eq:24} 
\bm{z}_j^\kappa =\ ^G\bm{u}_j^T\ (^G\bm{p}_j - ^G\bm{q}_j)
\end{equation}

$r_j$ in Equation (\ref{eq:23}) incorporates the post-undistortion uncertainty and can thus be treated as residual weighting:
\begin{equation}  \label{eq:25} 
r_j=\ ^G\bm{u}_j^T\ ^G\bm{n}_j \sim N(0, \bm{R}_j), \quad 
 \bm{R}_j=\ ^G\bm{u}_j^T\ ^G\bm{\Sigma}_{\bm{p}_j}\ ^G\bm{u}_j
\end{equation}

Finally, updating the state $\bm{x}$ is equivalent to updating the error state $\bm{\tilde{x}}_{k}^{\kappa}$. We solve the following equation:
\begin{equation} \label{eq:26} 
\min_{\bm{\tilde{x}}_{k}^{\kappa}}(\lVert \bm{x_k}\boxminus\bm{\Hat{x}}_k \rVert ^2_{\bm{\Hat{P}}_k} + \sum_{j} \lVert \bm{z}_j^\kappa + \bm{H}_j^\kappa \bm{\tilde{x}}_{k}^{\kappa} \rVert ^2_{\bm{R}_j})
\end{equation} 
\noindent where $\boxminus$ denotes the on-manifold minus. $|\bm{x}_k\boxminus\bm{\Hat{x}}_k|$ originates from the forward propagation, and $\bm{\Hat{P}}_k$ is the propagated state covariance. The second term constrains the state based on residuals from all valid measurements. This MAP problem can then be solved in an Kalman Filter framework:
\begin{align}   \label{eq:27}
\begin{split}
    \bm{K} &= (\bm{H}^T \bm{R}^{-1} \bm{H} + \bm{P}^{-1})^{-1} \bm{H}^T \bm{R}^{-1} \\
    \bm{\hat{x}}_k^{\kappa+1} &= \bm{\hat{x}}_k^{\kappa} \boxplus (-\bm{K}\bm{z}_k^\kappa - (\bm{I}-\bm{KH})(\bm{J}^\kappa)^{-1}(\bm{\hat{x}}_k^{\kappa} \boxminus \bm{\hat{x}}_k))
\end{split}
\end{align} 
\noindent where $\bm{H}=[\bm{H}_1^\kappa, ..., \bm{H}_M^\kappa]^T, \bm{R}=[R_1^\kappa, ..., R_M^\kappa]$ and $\bm{z}_k^\kappa = [(\bm{z}_1^\kappa)^T, ..., (\bm{z}_M^\kappa)^T]$ are integrated variables from all valid points in single scan, $\bm{J}^\kappa$ is defined in \cite{Fastlio2}.




\section{Experiments}   \label{sec:4}

\subsection{The 3DoF Vibration Platform Experiment}

\begin{figure*}[th]
  \centering
  \includegraphics[width=0.9\linewidth]{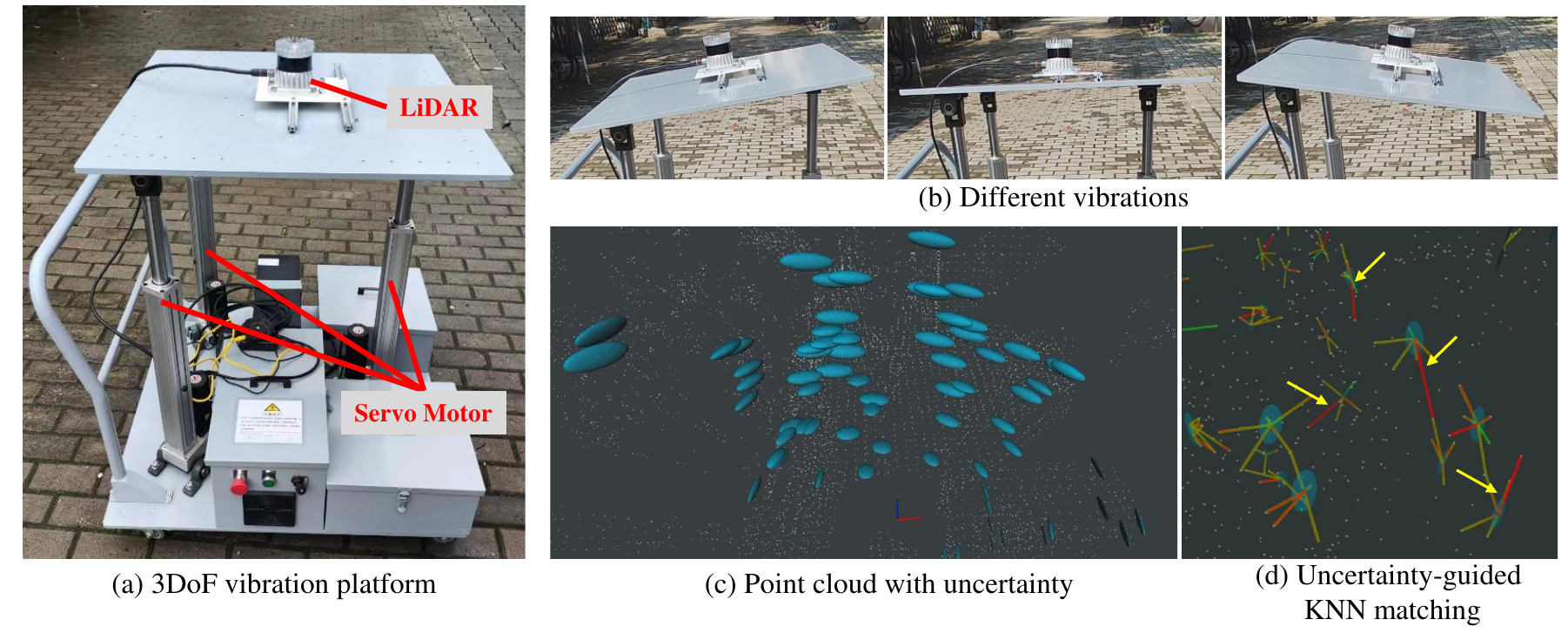}
  \caption{3DoF vibration platform experiments. (a) The 3DoF vibration platform. (b) Different vibrations of the platform. (c) and {\color{black}(d)} show LiDAR scans with post-undistortion uncertainty and matching results. The black ellipsoids represent post-undistortion uncertainties, while the white dots correspond to the kd-tree. Lines in (d) indicate KNN matching with $K$=5. Green lines represent original KNN matching, whereas red lines denote uncertainty-guided KNN matching. When the two coincide, the lines appear orange. The yellow arrows highlight the differences.}
  \label{Fig:3dofVisual}
\end{figure*}

To validate the robustness of the algorithm under severe vibration conditions, we conducted experiments using a 3DoF vibration platform. The platform consists of three high-precision electric cylinders, each with an effective stroke of 40 cm and a maximum speed of 33 cm/s. The electric cylinders adjust the planar surface at the top of the platform (Fig.\ref{Fig:3dofVisual}(a)(b)), achieving a static position accuracy error of less than 0.1 mm and an angular error of less than 0.5$^\circ$. 

The experimental procedure involved mounting the LiDAR on the platform and subjecting it to various vibrations for 30-second intervals. The platform returned to its initial position and remained stationary after vibration. We measured the end-time translation and rotation error, defined as the deviation between the algorithm's estimated translation/rotation after stabilization and the initial reference frame.

\begin{table*}[]
\centering
\caption{Average and standard deviation of end-time translation error (cm) and rotation error (deg) for Fast-LIO\cite{Fastlio2} and our method in 3DoF vibration platform experiments.}
\label{Tab:3dof}
\begin{tabular}{c|cc|cc}
\hline
\multirow{2}{*}{Vibration Descirption} & \multicolumn{2}{c|}{End-time Translation Error (cm)} & \multicolumn{2}{c}{End-time Rotation Error (deg)} \\ \cline{2-5} 
                                       & Fast-LIO\cite{Fastlio2}        & Ours           & Fast-LIO      & Ours         \\ \hline 
1 Hz z-axis linear vib.                 & 2.19 $\pm$ \textbf{0.28}       & \textbf{1.72} $\pm$ 0.81      & 0.049 $\pm$ 0.045   & \textbf{0.047} $\pm$ \textbf{0.036}  \\
2 Hz y-axis(pitch) angular vib.                 & 4.00 $\pm$ 0.79       & \textbf{3.75} $\pm$ \textbf{0.46}      & 0.194 $\pm$ \textbf{0.077}   & \textbf{0.188} $\pm$ 0.089  \\
3 Hz x-axis(roll) angular vib.                  & 3.18 $\pm$ \textbf{0.39}       & \textbf{2.91} $\pm$ 0.57      & 0.148 $\pm$ \textbf{0.024}   & \textbf{0.139} $\pm$ 0.038  \\
Hybrid linear \& angular vib.          & 4.75 $\pm$ \textbf{0.54}       & \textbf{4.28} $\pm$ 0.74      & 0.196 $\pm$ \textbf{0.105}   & \textbf{0.168} $\pm$ 0.107  \\ \hline
Average                                & 3.53 $\pm$ \textbf{0.50}       & \textbf{3.16} $\pm$ 0.64      & 0.145 $\pm$ \textbf{0.063}   & \textbf{0.135} $\pm$ 0.068  \\ \hline
\end{tabular}
\end{table*}

\begin{figure}[t]
  \centering
  \includegraphics[width=0.98\linewidth]{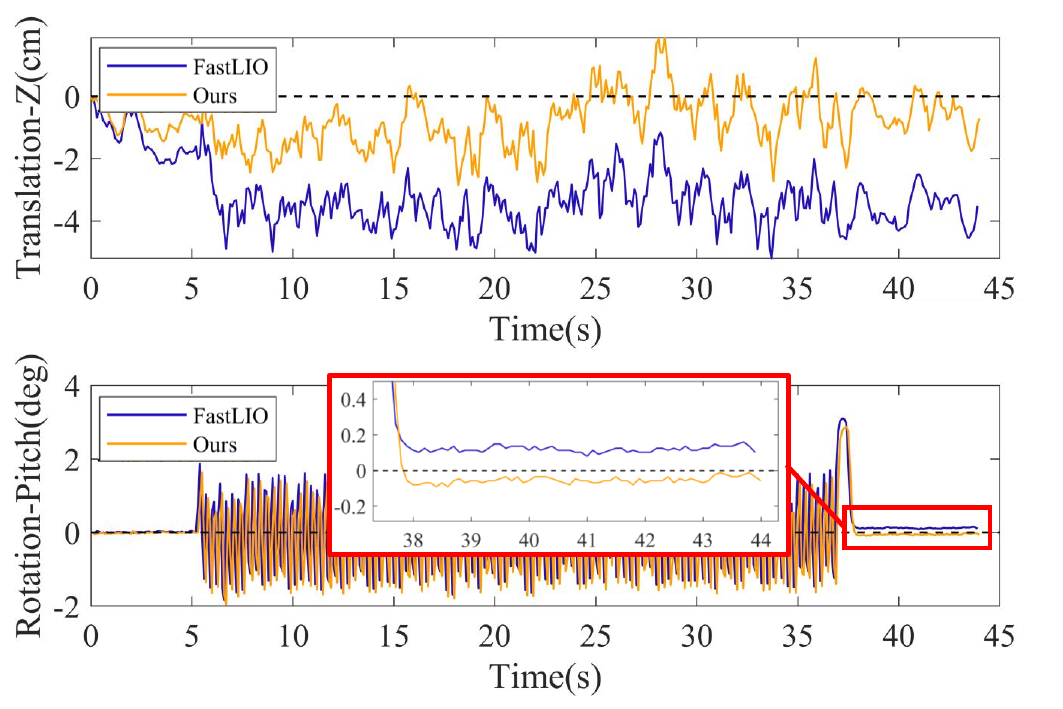}
  \caption{The estimated motion of Fast-LIO and our method in the vibration table experiment. Top: translation along the z-axis. Bottom: rotation along the pitch axis. Our method remains closer to the initial position after the vibration subsides.}
  \label{Fig:3dof_plot}
\end{figure}

Four experimental configurations were tested: (1) 1 Hz vertical linear vibration, (2) 2 Hz pitch angular vibration, (3) 3 Hz roll angular vibration, and (4) a hybrid motion incorporating all three types of vibration. Figure \ref{Fig:3dofVisual}(c) presents the point cloud with post-undistortion uncertainty, while Fig.\ref{Fig:3dofVisual}(d) displays the uncertainty-guided matching results. We conducted 5 trails for each vibration type, and the numerical end-time translation/rotation errors and corresponding standard deviations are summarized in Table \ref{Tab:3dof}. Our method has lower translation and rotation errors, but slightly larger standard deviations, {\color{black}which stems from the computational burden in Mahalanobis distance calculation and introduces additional uncertainty to the ikdtree \cite{ikdtree} updates.} Figure \ref{Fig:3dof_plot} illustrates the displacement along the z-axis and the pitch angle during the 3 Hz pitch motion, demonstrating that our method effectively maintains the z-axis position near zero and achieves superior pitch angle accuracy upon motion cessation.

\subsection{Public Dataset Experiments} \label{sec:4.1}

\textit{Dataset Selection:} To evaluate the algorithm's performance on intense vibration situations, we select several sequences from public datasets. NCD \cite{NCD} is a dataset collected by a handheld device equipped with an Ouster OS1-64 LiDAR. We use the \textit{05 quad-with-dynamics} and \textit{06 dynamic-spinning} sequence, which involve relatively intense motion. M2DGR \cite{M2DGR} is a multi-modal and multi-scenario ground robot dataset that uses the Velodyne VLP-32C LiDAR. We choose outdoor high-speed sequences \textit{street\_03} and \textit{street\_08}. The Botanic Garden Dataset \cite{BotanicGarden} is designed for challenging unstructured environments, where a ground robot moves over gravel, grass, and other terrains, experiencing significant vibration. It includes a Velodyne VLP-16C LiDAR, and we selected the \textit{1008-00} and \textit{1008-13} segments.

We compare the proposed method with LOAM \cite{LOAM}, LIO-SAM \cite{LIO-SAM}, LIO-Mapping \cite{Lio-Mapping}, Fast-LIO \cite{Fastlio2} and Point-LIO \cite{PLIO}, evaluating odometry performance using the mean value and root mean square error (RMSE) of absolute pose error (APE). Our method outperforms others except Botanic Garden \textit{1008-13}, where it ranks second. Detailed results are presented in Table \ref{Tab1:Compare}.

\subsection{The Uneven Terrain Experiment}  \label{sec:4.2}

\begin{figure*}[t]
  \centering
  \includegraphics[width=0.99\linewidth]{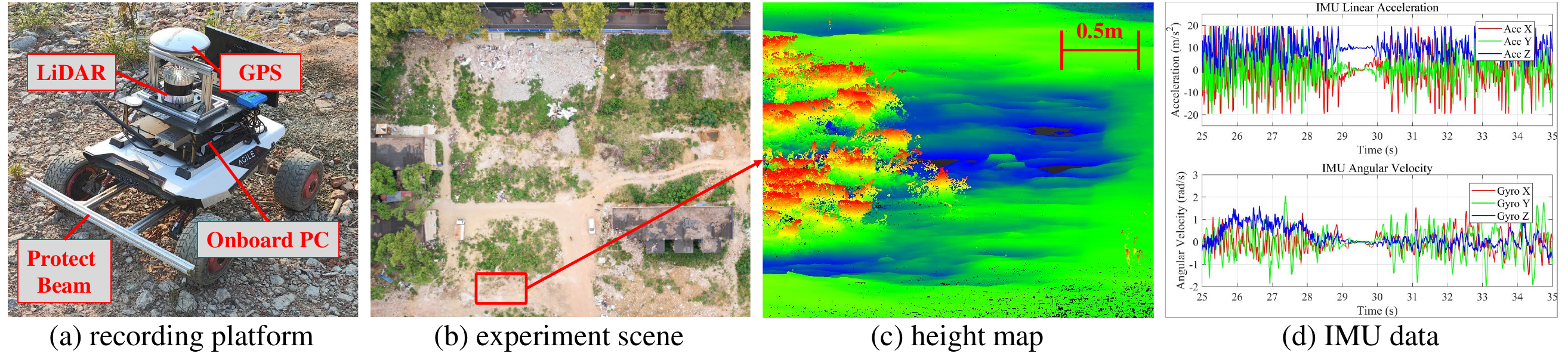}
  \caption{The platform and scene in our uneven terrain experiment. (a) Our wheeled platform. (b) A bird-eye view of the whole scene. (c) The elevation map scanned by a survey-grade scanner, where local height variations can reach 5-10 cm, causing significant bumps to our platform's wheel with a diameter of 17cm. (d) IMU data recorded during the platform's operation, showing significant variations in acceleration and angular velocity.}
  \label{Fig:Car}
\end{figure*}

\begin{table*}[th]
\centering
\caption{Mean/RMSE APE (m)  of different LIO methods on public dataset sequences}
\label{Tab1:Compare}
\resizebox{0.8\textwidth}{!}{%
\begin{tabular}{ccccccc}
\hline
\multicolumn{1}{c|}{\multirow{2}{*}{Method}} & \multicolumn{2}{c|}{NCD \cite{NCD}}                                         & \multicolumn{2}{c|}{M2DGR \cite{M2DGR}}                                       & \multicolumn{2}{c}{Botanic Garden \cite{BotanicGarden}}          \\
\multicolumn{1}{c|}{}                        & {05}                   & \multicolumn{1}{c|}{06}                   & {street03}             & \multicolumn{1}{c|}{street08}             & {1018-00}              & {1018-13}              \\ \hline
\multicolumn{1}{c|}{LOAM \cite{LOAM}}                    & 0.305/0.334          & \multicolumn{1}{c|}{2.642/3.594}          & 0.745/0.859          & \multicolumn{1}{c|}{2.123/2.908}          & 0.478/0.052          & 0.128/0.151          \\
\multicolumn{1}{c|}{LIO-SAM \cite{LIO-SAM}}                 & F(10)$^*$               & \multicolumn{1}{c|}{0.206/0.241}          & F(155)$^*$               & \multicolumn{1}{c|}{F(80)$^*$}                & 0.072/0.088          & 0.078/0.090          \\
\multicolumn{1}{c|}{LIO-Mapping \cite{Lio-Mapping}}             & 0.148/0.180          & \multicolumn{1}{c|}{0.681/0.771}          & 0.390/0.511          & \multicolumn{1}{c|}{0.179/0.206}          & 0.065/0.071          & 0.247/0.350          \\
\multicolumn{1}{c|}{Fast-LIO \cite{Fastlio2}}               & 0.097/0.147          & \multicolumn{1}{c|}{0.078/0.090}          & 0.112/0.128          & \multicolumn{1}{c|}{0.156/0.175}          & 0.034/0.039          & \textbf{0.054/0.059}          \\
\multicolumn{1}{c|}{Point-LIO \cite{PLIO}}               & 0.224/0.247          & \multicolumn{1}{c|}{0.296/0.380}          & 0.169/0.180          & \multicolumn{1}{c|}{0.187/0.216}          & 0.038/0.043          & 0.069/0.076          \\
\multicolumn{1}{c|}{Our method}                    & \textbf{0.066/0.083} & \multicolumn{1}{c|}{\textbf{0.065/0.081}} & \textbf{0.111/0.123} & \multicolumn{1}{c|}{\textbf{0.135/0.153}} & \textbf{0.029/0.034} &  {0.064/0.075} \\ \hline
\multicolumn{7}{l}{* F($s$): fails to give a roughly correct trajectory after $s$ second.}                                                                                                                   
\end{tabular}%
}
\end{table*}

\textit{The Experiment Platform and Scene:} We set up a wheeled robot equipped with an Ouster OS1-32 LiDAR. To obtain ground truth for the trajectory, we mounted an RTK module (Harxon TS101) with centimeter-level horizontal positioning accuracy. Since RTK exhibits slightly larger measurement errors in elevation (LiDAR's z-axis), we evaluate APE only on the 2D plane. The experimental platform and scene are shown in Fig.\ref{Fig:Car}. The platform experience intensive vibration during the operation due to the uneven terrains, with the acceleration and angular velocity up to 20 $m/s^2$ and 2 $rad/s$.

We collect several sequences:
\textit{01 circle} is robot moving around the field, with an average speed about 0.5m/s;
\textit{02 sharp-turn} includes several sharp-turns, with a maximum turning angular velocity of 270°/s;
\textit{03 collision}, the robot collides multiple times with obstacles such as tree trunks and rocks in the field;
\textit{04 all-challenges} includes fast movements, hard braking and sudden acceleration, sharp turns, and collisions.

\begin{figure}[t]
  \centering
  \includegraphics[width=0.98\linewidth]{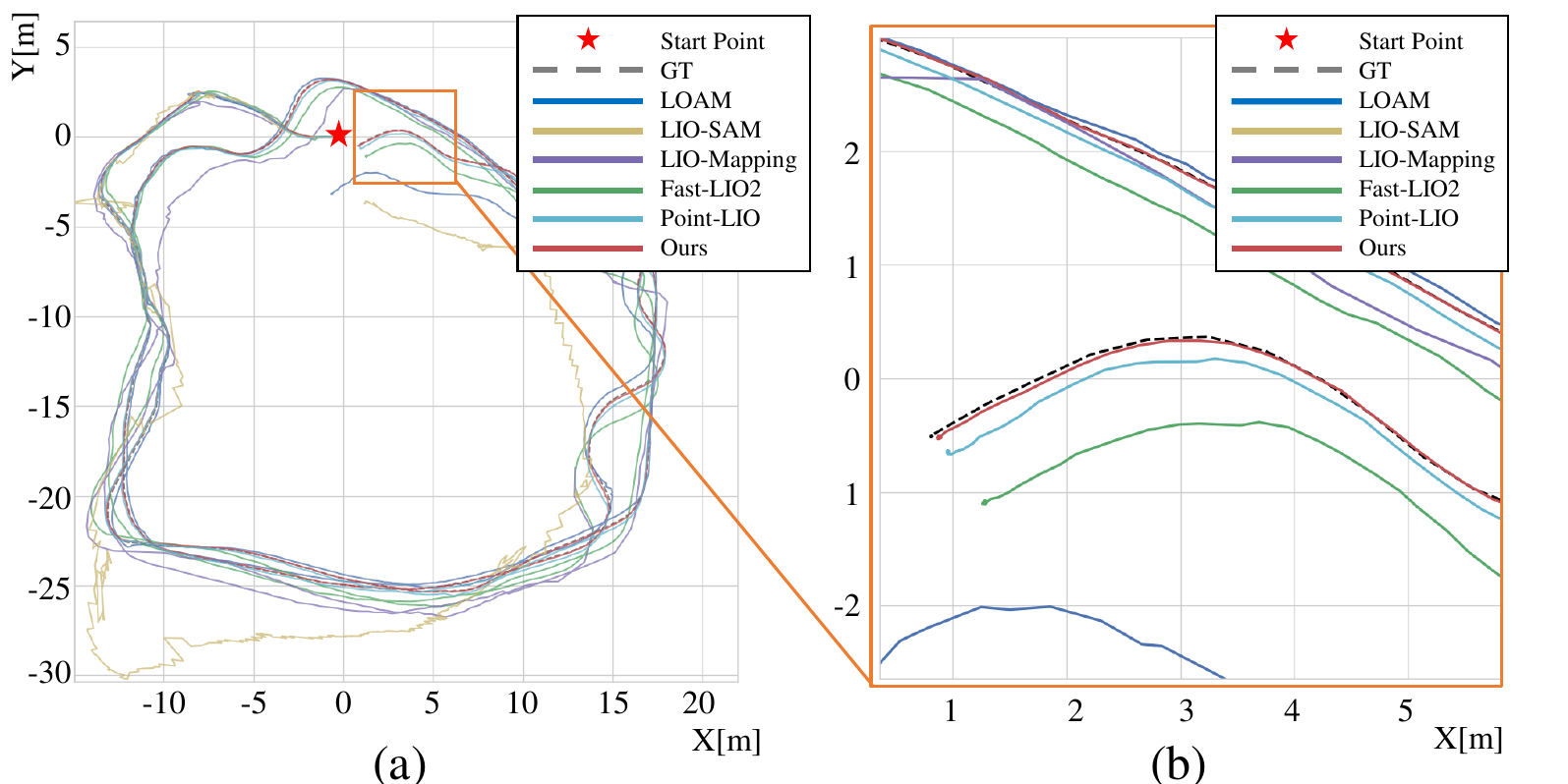}
  \caption{Trajectories of different methods on our \textit{02 sudden-turn} sequence. (a) The whole view. (b) Zoomed at the trajectory end.}
  \label{Fig7:trajectoryCompare}
\end{figure}

\begin{figure}[t]
  \centering
  \includegraphics[width=0.95\linewidth]{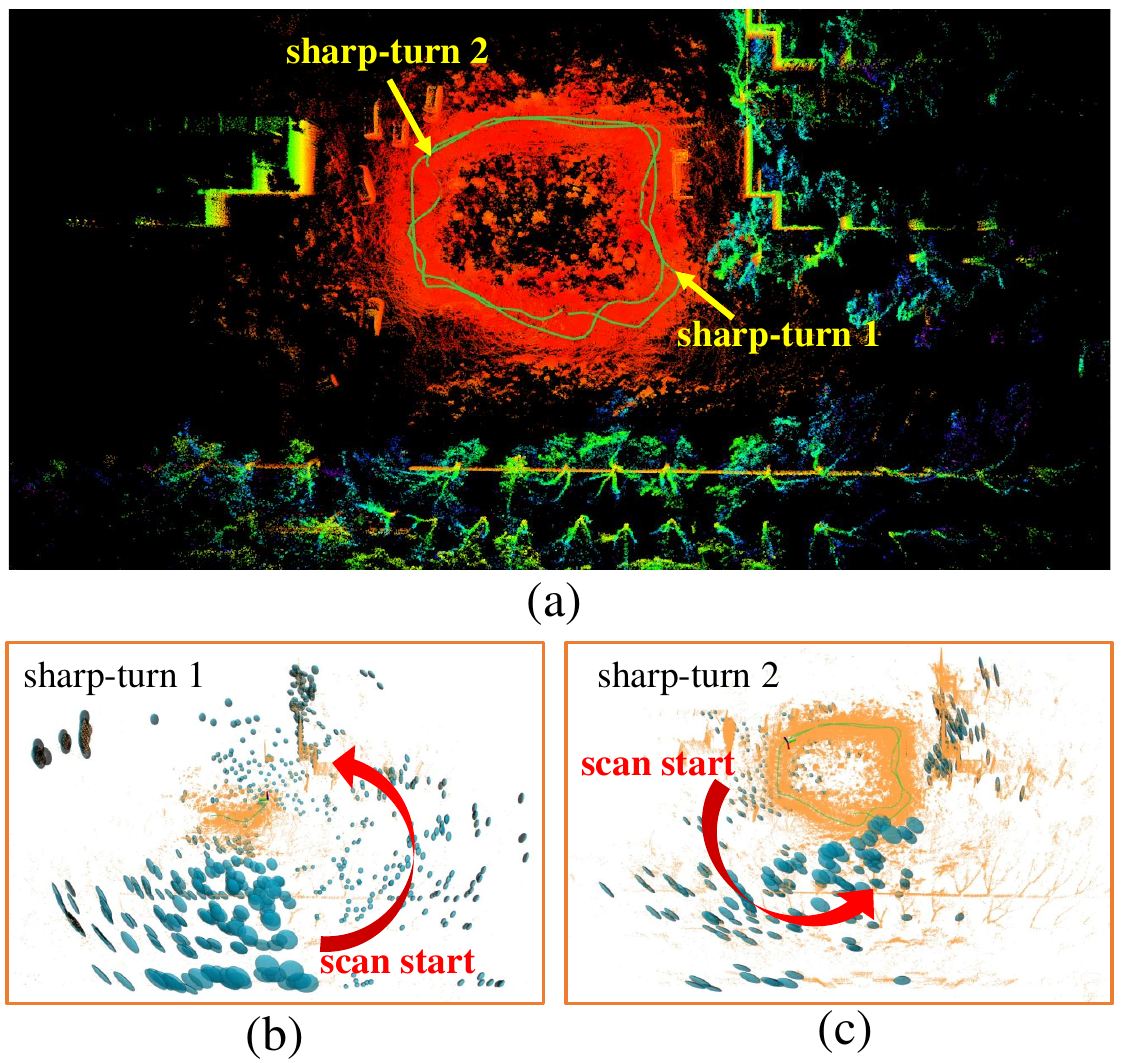}
  \caption{The execution process of the proposed method. (a) The trajectory and the 3D map on \textit{02 sharp-turn}. The map color is encoded by the elevation. {\color{black}(b)(c)} Two scans with post-undistortion uncertainty (black ellipsoids) during sharp-turns in (a). {\color{black}The red arrow shows the scan direction, and points sampled at the end of the scan have larger post-undistortion uncertainties compared to points sampled at the beginning.}}
  \label{Fig8:running}
\end{figure}

The quantitative experimental results are shown in Table \ref{Tab2:Own} showing that the proposed algorithm achieves the best performance on all 4 sequences. We visualize the trajectories of different methods alongside the ground truth in Fig.\ref{Fig7:trajectoryCompare}. In Fig.\ref{Fig8:running}, we show the trajectory and mapping of the proposed algorithm on \textit{02 sharp-turn} as well as the scans with post-undistortion uncertainties at two sharp-turn moments.

\begin{table}[]
\centering
\caption{MEAN APE(m) of different LIO methods on own recordings}
\label{Tab2:Own}
\resizebox{\linewidth}{!}{%
\begin{tabularx}{\linewidth}{>{\centering\arraybackslash}p{2.3cm}| >{\centering\arraybackslash}p{0.8cm} >{\centering\arraybackslash}p{1.2cm} >{\centering\arraybackslash}p{1cm} >{\centering\arraybackslash}p{1.8cm}}
\hline
\multirow{2}{*}{Method}         & 01 \quad circle         & 02 sharp-turn    & 03 collision     & 04 \quad all-challenges   \\ \hline
\multicolumn{1}{c|}{LOAM \cite{LOAM}}           & 0.082             & 2.103            & 6.452            & 2.064               \\
\multicolumn{1}{c|}{LIO-SAM \cite{LIO-SAM}}        & {0.042}             & F(55)$^*$   & F(25)$^*$   & F(110)$^*$     \\
\multicolumn{1}{c|}{LIO-Mapping \cite{Lio-Mapping}}    & 0.070             & 6.095            & 40.162           & 12.918              \\
\multicolumn{1}{c|}{Fast-LIO \cite{Fastlio2}}       & 0.047             & 0.097            & 0.073            & 0.062               \\
\multicolumn{1}{c|}{Point-LIO \cite{PLIO}}      & 0.045             & {0.064}            & {0.059}            & {0.060}               \\
\multicolumn{1}{c|}{Our method}           & \textbf{0.026}    & \textbf{0.038}   & \textbf{0.056}   & \textbf{0.057}      \\ \hline
\multicolumn{5}{l}{\scriptsize* F($s$): fails to give a roughly correct trajectory after $s$ second.}
\end{tabularx}%
}
\end{table}

\subsection{Computational Complexity Evaluation}  \label{sec:4.3}

We implemented the proposed method in ROS using the C++ APIs and tested its computational performance on a 3.8GHz AMD CPU mini-PC. We downsample each scan to 1/4, resulting in approximately 2,500 points, and set the following parameters: 0.5 m for ikdtree resolution, 5 for the KNN search count, and 4 as the maximum iteration count for iKF, {\color{black}which are the general settings for all experiments in this letter. For the M2DGR sequences,} The most time-consuming steps of the algorithm are calculating the uncertainty for each point (around 6ms) and uncertainty-guided matching (needing 2-3 iterations per scan and taking a total of 22 ms on average). The total processing time per scan is approximately 36 ms, thus meeting real-time processing requirements.

\begin{table}[]
\centering
\caption{{\color{black}Ablation Study of the proposed method. UM: post-undistortion uncertainty modeling. GM: uncertainty-guided matching. HM: heuristic method for vibration intensity estimation.}}
\label{Tab4:AblationStudy}
\begingroup
\color{black}
\begin{tabular}{c|ccc|c}
\hline
Setting                                & UM  & GM  & HM & mean APE \\ \hline
w/o uncertainty \& guided matching                        & $\times$ & $\times$ & MAD              & 0.042    \\
w/o uncertainty                        & $\times$ & \checkmark & MAD              & 0.039    \\
w/o guided matching                    & \checkmark & $\times$ & MAD              & 0.031    \\
vib. estimation by LLS  & \checkmark & \checkmark & LLS              & 0.028    \\
vib. estimation by STD & \checkmark & \checkmark & STD              & 0.026    \\ 
\textbf{Proposed}                               & \checkmark & \checkmark & MAD              & \textbf{0.026}    \\ \hline
\end{tabular}
\endgroup
\end{table}

{\color{black}
\subsection{Ablation Study and Different Vibration Estimation Methods} \label{sec:4.4}

To validate the effectiveness of the post-undistortion uncertainty modeling (UM) and uncertainty-guided matching (GM), we conducted ablation experiments. Without UM or GM modules, mean APE increases considerably. In Table \ref{Tab4:AblationStudy} we show the numeric results on our \textit{01 circle} sequence. 

In Equation (\ref{eq:16}), we adopt mean absolute deviation (MAD) as the default heuristic method (HM) for estimating vibration intensity. We compare it with standard deviation (STD) and linear least squares error (LLS), two common alternatives for measuring variation. Mean APEs on \textit{01 circle} are also reported in Table \ref{Tab4:AblationStudy}, and all heuristics yield similar performance. MAD is adopted in our method because of its calculation simplicity.
}

\section{Conclusion}

In this paper, we first highlight the challenges of achieving accurate point cloud undistortion under intense vibration conditions is difficult considering the constraints of computational resources and hardware costs. To address this, we introduce point-wise post-undistortion uncertainty to model the positional uncertainty of each point after the undistortion process. The proposed method formulates the undistortion error induced by angular and linear vibrations using a Gaussian distribution, incorporating rotational, translational, and measurement uncertainties for each point. This uncertainty further guides point-to-map matching and adjusts residual weighting in an iterated Kalman filter (iKF) to refine odometry estimation. We conducted vibration platform experiments, demonstrating that the proposed odometry method achieves lower cumulative errors under vibrational conditions. Additionally, experiments on both public datasets and our own mobile platform confirm that the proposed uncertainty modeling effectively enhances trajectory tracking accuracy. We also released our recorded sequences, which capture unstructured terrains with intense vibrations, to facilitate further research.


\bibliographystyle{IEEEtran}
\bibliography{IEEEabrv, ref}

\vfill

\end{document}